\pdfoutput=1

\documentclass[11pt,dvipsnames]{article}

\usepackage{acl}

\usepackage{graphics}
\usepackage[table]{xcolor}
\usepackage{etoolbox}
\usepackage{pgf}
\usepackage{multirow}
\usepackage{tabularx,stackengine}

\newcounter{notecounter}
\newcommand{\enotesoff}{\long\gdef\enote##1##2{}}
\newcommand{\enoteson}{\long\gdef\enote##1##2{{
			\stepcounter{notecounter}
			\large\bf
			\hspace{100cm}\arabic{notecounter} $<<<$ ##1: ##2
			$>>>$\hspace{1cm}}}}
\enoteson
\enotesoff

\usepackage{times}
\usepackage{latexsym}
\usepackage{tikz}
\usetikzlibrary{shapes,shapes.multipart}
\usetikzlibrary{positioning}
\usetikzlibrary{fit}
\usepackage{adjustbox}
\usepackage{amssymb}
\usepackage{booktabs}

\usepackage[T1]{fontenc}

\usepackage[utf8]{inputenc}

\usepackage{microtype}

\title{Why don't people use character-level machine translation?}

\author{Jindřich Libovický$^1$ \and Helmut Schmid$^2$ \and Alexander Fraser$^2$ \\
  $^1$ Faculty of Mathematics and Physics, Charles Univeristy, Prague, Czech Republic \\
  $^2$ Center for Information and Speech Processing, LMU Munich, Germany \\
  \texttt{libovicky@ufal.mff.cuni.cz \{schmid, fraser\}@cis.lmu.de}}

\begin{document}
\maketitle
\begin{abstract}
We present a literature and empirical survey that critically assesses the state
    of the art in character-level modeling for machine translation (MT).
Despite evidence in the literature that character-level systems are comparable
    with subword systems, they are virtually never used in competitive setups
    in WMT competitions.
We empirically show that even with recent modeling innovations in
    character-level natural language processing, character-level MT systems
    still struggle to match their subword-based counterparts.
Character-level MT systems show neither better domain robustness, nor better
    morphological generalization, despite being often so motivated.
        %
However, we are able to show 
robustness towards source side noise and
that
    translation quality does not degrade with increasing beam size at decoding
    time.
\end{abstract}

\section{Introduction}

The progress in natural language processing (NLP) brought by deep learning is
often narrated as removing assumptions about the input data and letting the
models learn everything end-to-end. One of the assumptions about input data
that seems to resist this trend is (at least partially) linguistically
motivated segmentation of input data in machine translation (MT) and NLP in
general.

For NMT, several papers have claimed parity of character-based methods with
subword models, highlighting advantageous features of such systems.
Very recent examples include
\citet{gao-etal-2020-character,banar2020character,li-etal-2021-char}. Despite
this, character-level methods are rarely used as strong baselines in research
papers and shared task submissions, suggesting that character-level models
might have drawbacks that are not sufficiently addressed in the literature.

In this paper, we examine what the state of the art in character-level MT
really is. We survey existing methods and conduct a meta-analysis of the input
segmentation methods used in WMT shared task submissions.
We then systematically compare the most recent character-processing
architectures, some of them taken from general NLP research and used for the
first time in MT\@.
\enote{AF}{This sounds a bit like we will *only* use the two-step decoder,
maybe reword slightly?}
Further, we propose an alternative two-step decoder architecture that unlike
standard decoders does not suffer from a slow-down due to the length of
character sequences.
Following the recent findings on MT decoding, we evaluate different decoding
strategies in the character-level context.

Many previous studies on character-level MT drew their conclusions from
experiments on rather small datasets and focused only on quantitatively
assessed translation quality without further analysis. To compensate for this,
we revisit and systematically evaluate the state-of-the-art approaches to
character-level neural MT and identify their major strengths and weaknesses on
large datasets.


\section{Character-Level Neural MT}

%
Character-level processing was hardly possible within the statistical MT
paradigm that assumed the existence of phrases consisting of semantically rich
tokens that roughly correspond to words. Neural sequence-to-sequence models
\citep{sutskever2014sequence,bahdanau2015neural,vaswani2017attention} do not
explicitly work with this assumption. In theory, they can learn to transform
any sequence into any sequence.

The original sequence-to-sequence models used word-based vocabularies of a
limited size and which led to a relatively frequent occurrence of
out-of-vocabulary tokens. A typical solution to that problem is subword
segmentation
\citep{sennrich-etal-2016-neural,kudo-richardson-2018-sentencepiece}, which
keeps frequent tokens intact and splits less frequent ones into smaller units.


Modeling language on the character level is attractive because it can help
overcome several problems of subword models. One-hot representations of words or
subwords do not reflect systematic character-level relations between words,
potentially harming morphologically rich languages. With subwords, minor typos
on the source side lead to radically different input representations resulting
in low robustness towards source-side noise
\citep{provilkov-etal-2020-bpe,libovicky-fraser-2020-towards}.

Models using recurrent neural networks (RNNs) showed early success with
character-level segmentation on the decoder side
\citep{chung-etal-2016-character}. Using character-level processing on the
encoder side proved harder which was attributed to the features of the
attention mechanism which can presumably benefit from semantically rich units
(such as subwords) in the encoder. Following this line of thinking,
\citet{lee-etal-2017-fully} introduced 1D convolutions with max-pooling that
pre-process the character sequence into a sequence of latent word-like states.
%
%
Coupled with a character-level decoder, they claimed to match the
state-of-the-art subword-based models. Even though this architecture works well
on the character level, it does not generalize further to the byte level
\citep{costa-jussa-etal-2017-byte}. Hybrid approaches combining tokenization
into words 
with 
the computation of character-based word representations
were successfully used with RNNs
\citep{luong-manning-2016-achieving,gronroos-etal-2017-extending,ataman-etal-2019-importance}.
Later, \citet{cherry-etal-2018-revisiting} showed that 
RNNs perform on par with subword models without changing the model architecture if the models are sufficiently large. 
\citet{kreutzer-sokolov-2018-learning} support this by showing
that RNN models which learn segmentation jointly with the rest of the model are
close to character-level.

Character-level modeling with Transformers appears to be more difficult.
\citet{gupta2019characterbased} used Transparent Attention
\citep{bapna-etal-2018-training} to train deep character-level models and
needed up to 32 layers to close the gap between the BPE and character models,
which makes the model too large for practical use.
\citet{libovicky-fraser-2020-towards} narrowed the gap between subword and
character modeling using curriculum learning by finetuning subword models to
character-level.

\citet{gao-etal-2020-character} proposed adding a convolutional sub-layer in
the Transformer layers. At the cost of a 30\% increase in parameter count, they
managed to narrow the gap between subword- and character-based models by half.
\citet{banar2020character} reused the convolutional preprocessing layer with
constant-size segments 
of \citet{lee-etal-2017-fully} in a Transformer model
for translation into English. 
Without changing the decoder, 
they reached comparable, but usually slightly worse, translation quality 
compared to 
BPE-based models.

\begin{figure}
    \centering
    \includegraphics{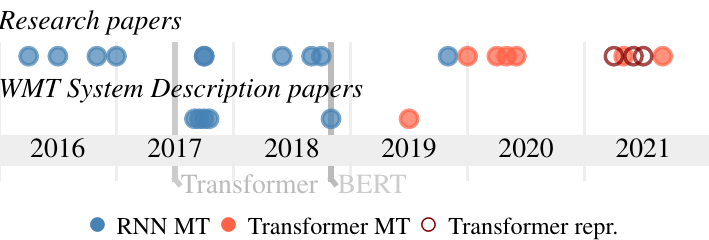}


    \caption{A timeline of research interest in character-level MT\@. Months of
    arXiv pre-print publication of the papers cited in Sections~2 and~3.
    Transformer repr.\ means pre-trained general-purpose sentence
    representation, not MT models.}\label{fig:timeline}

\end{figure}

\citet{shaham-levy-2021-neural} revisited character- and byte-level MT on
rather small IWSLT datasets. Their results show that 
character-level and byte-level models are usually worse than BPE models, 
but 
byte-based models without embedding layers often outperform BPE-based models in the
out-of-English direction. Using similarly small datasets,
\citet{li-etal-2021-char} claim that character-level modeling outperforms BPE
when translating into fusional, agglutinative, and introflexive languages.

%
\citet{nikolov-etal-2018-character} experimented with character-level models
for romanized Chinese.
These models performed comparable to models 
using logographic signs,
but 
significantly worse than
models 
using subwords.
\citet{zhang-komachi-2018-neural} argued that signs in logographic languages
carry too much information and were able to improve the translation quality by
segmenting Chinese and Japanese into sub-character units while keeping subword
segmentation on the English side.

Little is known about other properties of character-level MT beyond the overall
translation quality. \citet{sennrich-2017-grammatical} prepared a set of
contrastive English-German sentence pairs and tested them using shallow
RNN-based models. They observed that character-based models transliterated
better, but captured morphosyntactic agreement worse.
\citet{libovicky-fraser-2020-towards} evaluated Transformer-based
character-level models using MorphEval and came to mixed conclusions.

\citet{gupta2019characterbased} and \citet{libovicky-fraser-2020-towards} make
claims about the noise robustness of the character-level models using synthetic
noise. \citet{li-etal-2021-char} evaluated domain robustness by training models
on small domain-specific datasets and evaluating them on unrelated domains,
claiming the superiority of character-level models in this setup.
On the other hand, \citet{gupta2019characterbased} evaluated the domain
robustness in a more natural setup and did not observe higher robustness when
evaluating general domain models on domain-specific tests compared to BPE\@.

Another consideration is longer training and inference times.
Character-level systems are significantly slower due to the increased sequence length.
\citet{libovicky-fraser-2020-towards} reported a 5.6-fold 
slowdown at training time and a 4.7-fold 
slowdown at inference time compared to subword models.

Recent research on character-level modeling goes beyond MT\@. Pre-trained
multilingual representations are a particularly active area.
\citet{clark2021canine} propose CANINE\@. The model shrinks character sequences
into fewer hidden states (similar to \citealp{lee-etal-2017-fully}). They use
local self-attention and strided convolutions (instead of highway layers and
max-pooling as in Lee's work). Their model is either trained using the
masked-language-modeling objective \citep{devlin-etal-2019-bert} with subword
supervision, or in an encoder-decoder setup similar to
\citet{raffel-etal-2020-exploring}. Both 
methods 
reach a representation quality comparable to similar subword models.

ByT5 \citep{xue2021byt5} and Charformer \citep{tay2021charformer} are based on
the mT5 model \citep{xue-etal-2021-mt5} which uses sequence-to-sequence
denoising pre-training. Whereas byT5 only uses byte sequences instead of
subwords and differs in hyperparameters, Charformer uses convolution and
combines character blocks to obtain latent subword representations.  These
models mostly reach similar results to sub-word models, occasionally
outperforming a few of them, in the case of Charformer without a significant
slowdown.
\enote{AF}{This writing tends to make me think that we are saying Charformer is
good, and therefore we will use it.}

\section{WMT submissions}

The Conference on Machine Translation (WMT) organizes annual shared tasks in
various use cases of MT\@. The shared task submissions focus on translation
quality rather than the novelty of presented ideas, as most other research
papers do. Therefore, we assume that, if character-level models were a
fully-fledged alternative to subword models, at least some systems submitted to
the shared tasks would use character-level models.

\begin{figure}
    \centering
    \includegraphics{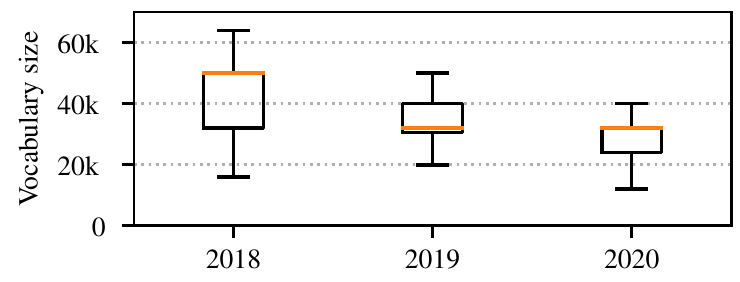}

    \caption{A boxplot of vocabulary sizes of WMT systems from 2018--2020, the
    median is denoted with the orange line.}\label{fig:vocabSize}

\end{figure}

We annotated recent system description papers with 
the 
input and output segmentation 
method 
they used. 
We focused on information about experiments with
character-level models. Since we are primarily interested in the Transformer
architecture that became the standard after 2017, we only included system
description papers from 2018--2020
\citep{bojar-etal-2018-findings,barrault-etal-2019-findings,barrault-etal-2020-findings}.
Transformers were used in 81\%, 87\%, and 97\% of the systems in the respective
years. We included the main task on WMT, news translation, and two minor tasks
where character-level methods might help: translation robustness
\citep{li-etal-2019-findings,specia-etal-2020-findings} and translation between
similar languages (ibid.).
%

Almost all systems use 
a 
subword-based vocabulary (BPE\@: 81\%, 71\%, 66\% in the
respective years; SentencePiece: None in 2018, 9\% and 25\% in the following
ones). Purely word-based (none in 2018, 2\% and 3\% in the later years) or
morphological segmentation (4\%, 2\%, 3\% in the respective years) are rarely
used. The average vocabulary size decreases over time (see
Figure~\ref{fig:vocabSize}) with a median size remaining at 32k in the last two
years. The reason for the decreasing average is probably a higher proportion of
systems for low-resource languages, where a smaller vocabulary leads to better
translation quality \citep{sennrich-zhang-2019-revisiting}.

Among the 145 annotated system description papers, there were only two that
used character-level segmentation. \citet{mahata-etal-2018-jucbnmt} used a
character-level model for Finnish-to-English translation. This system, however,
makes many suboptimal design choices and ended up as the last one in the manual
evaluation. \citet{scherrer-etal-2019-university} experimented with
character-level systems for similar language translation and observed that
characters outperform other segmentations for Spanish-Portuguese translation,
but not for Czech-Polish.
\citet{knowles-etal-2020-nrc} experimented with different subword vocabulary
sizes for English-Inuktikut translation and reached the best results using a
subword vocabulary of size 1k, which makes it close to the character level.
Most of the papers do not even mention character-level segmentation as a viable
alternative they would like to pursue in future work (7\% in 2018, 2\% in 2019,
none in 2020).

Character-level methods were more frequently used in WMT17 with RNN-based
systems, especially for translation of Finnish
\citep{escolano-etal-2017-talp,ostling-etal-2017-helsinki} and less
successfully for Chinese \citep{holtz-etal-2017-university} and the automatic
post-editing task \citep{varis-bojar-2017-cuni}.

On the other hand, Figure~\ref{fig:timeline} shows that the research interest
in character-level methods remains approximately the same, or may have slightly
increased. For practical solutions in WMT systems, we clearly show that system
designers in the WMT community have avoided character-level models.

We speculate that the main reasons for not considering character-level modeling
are its lower efficiency and the fact that the literature shows no clear
improvement of translation quality.
Most of the submissions use back-translation (85\%, 82\%, and 94\% in the
respective years), often iterated several times (11\%, 20\%, 16\%), which
requires both training and inference on large datasets. With the approximately
5-fold slowdown, WMT-scale experiments on character models are not easily
tractable.

\section{Evaluated Models}

We evaluate several Transformer-based architectures for character-level MT\@.
A major issue with character-level sequence processing is the sequence length
and low information density compared to subword sequences. Architectures for
character-level sequence processing typically address this issue by locally
processing and shrinking the sequences into latent word-like units. In our
experiments, we explore several ways to do this.

First, we directly use character embeddings as input to the Transformer.
Second, following \citet{banar2020character}, we use the convolutional
character processing layers proposed by \citet{lee-etal-2017-fully}. Third, we
replace the convolutions with local self-attention as proposed in the CANINE
model \citep{clark2021canine}. Finally, we use the recently proposed Charformer
architecture \citep{tay2021charformer}.

\paragraph{Lee-style encoding.} \citet{lee-etal-2017-fully} process the
sequence of character embeddings with convolutions of different kernel sizes
and number of output channels. In the original paper, this was followed by 4
highway layers \citep{srivastava2015highway}. In our preliminary experiments,
we observed that a too deep stack of highway layers leads to diminishing
gradients, and we replaced the second two Highway layers with feedforward
sublayers as used in the Transformer architecture \citep{vaswani2017attention}.

\paragraph{CANINE.} \citet{clark2021canine} experiment with character-level
pre-trained sentence representations. The character-processing architecture is
in principle similar to \citet{lee-etal-2017-fully} but uses more modern
building blocks. Character embeddings are processed by a Transformer layer with
local self-attention which only allows the states to attend to states in their
neighborhood. This is followed by downsampling using strided convolution.

Originally, CANINE used a local self-attention span as long as 128 characters.
In the case of MT, this would usually span the entire sentence, so we use
significantly shorter spans.

\paragraph{Charformer.} Unlike previous approaches, Charformer
\citep{tay2021charformer} does not apply a non-linearity on the embeddings and
gets latent subword representations by repeated averaging of character
embeddings.
First, it processes the sequence using a 1D convolution, so the states are
aware of their 
mutual local positions in local neighborhoods.
Second, non-overlapping character $n$-grams of length up to $N$ are represented
by averages of the respective character embeddings.
This means that for each character, there is a vector
that represents the character as a member of $n$-grams of length 1 to $N$.
In the third step, the character blocks are scored with a scoring function (a
linear transformation), which can be interpreted as attention over the $N$
different $n$-gram lengths. The attention scores are used to compute a weighted
average over the $n$-gram representations. Finally, the sequence is downsampled
using mean-pooling with window size and stride size $N$ (i.e., the maximum $n$-gram size).

Whereas Lee-style encoding allows using low-dimensional character embeddings
and keeps most parameters in the convolutional layers, CANINE and Charformer
need the character representation to have the same dimension as the following
Transformer layer stack.

\begin{figure}

        \centering
        \includegraphics{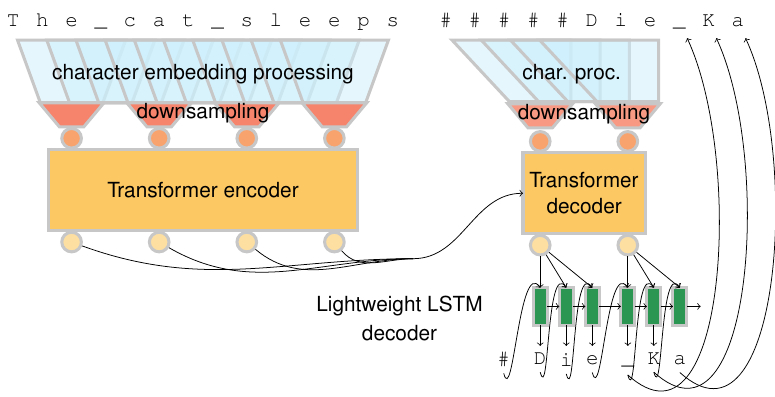}

        \caption{Encoder-decoder architecture with character-processing layers
        and a two-step decoder with lightweight LSTM for output
        coherence.}\label{fig:TwoStepDecoder}

\end{figure}

\paragraph{Two-step decoding.} The architectures mentioned above allow the
Transformer layers to operate more efficiently with a shorter and more
information-dense sequence of states. However, while decoding, we need to
generate the target character sequence in the original length, by outputting a
block of characters in each decoding step. Our preliminary experiments showed
that generating blocks of characters non-autoregressively leads to incoherent
output. Therefore, we propose a two-step decoding architecture where the stack
of Transformer layers operating over the downsampled sequence is followed by a
lightweight LSTM autoregressive decoder (see Figure~\ref{fig:TwoStepDecoder}).

The input to the LSTM decoder is a concatenation of the embedding of the
previously generated character and a projection of the Transformer decoder
output state. At inference time, the LSTM decoder generates a block of
characters and inputs them to the character-level processing
layer. The Transformer decoder computes an output state that the
LSTM decoder uses to generate another character block. More
details are in Appendix~\ref{sec:appendixTwoStep}.

Modifying Charformer for the two-step decoding would require a long padding at
the 
beginning 
of the sequence causing the decoder to diverge. Because of that, we
use Lee-style encoding on the decoder side when using Charformer in the
encoder.

First, we conduct all our experiments on the small IWSLT datasets. 
Then 
we evaluate the most promising architectures 
on larger datasets.

\section{Experiments on Small Data}

\begin{table*}
    \centering
    
\newcommand{\EnArBLEU}[2]{%
    \ifdimcomp{#1pt}{>}{13.5 pt}{\R{#1}{#2}}{%
    \ifdimcomp{#1pt}{<}{6.9 pt}{\R{#1}{#2}}{%
         \pgfmathparse{int(round(100*(#1/(13.5-6.9))-(6.9*(100/(13.5-6.9)))))}%
        \xdef\tempa{\pgfmathresult}%
        \cellcolor{high!\tempa!low!\opacity} \hspace{-8pt}\ifdimcomp{#1pt}{>}{0pt}{\phantom{-}}{}\R{#1}{#2}\hspace{-8pt}
    }}
 }

\newcommand{\EnArchrF}[2]{%
    \ifdimcomp{#1pt}{>}{0.448 pt}{\R{#1}{#2}}{%
    \ifdimcomp{#1pt}{<}{0.373 pt}{\R{#1}{#2}}{%
         \pgfmathparse{int(round(100*(#1/(0.448-0.373))-(0.373*(100/(0.448-0.373)))))}%
        \xdef\tempa{\pgfmathresult}%
        \cellcolor{high!\tempa!low!\opacity} \hspace{-8pt}\ifdimcomp{#1pt}{>}{0pt}{\phantom{-}}{}\R{#1}{#2}\hspace{-8pt}
    }}
 }

\newcommand{\EnArCOMET}[2]{%
    \ifdimcomp{#1pt}{>}{0.274 pt}{\R{#1}{#2}}{%
    \ifdimcomp{#1pt}{<}{-0.355 pt}{\R{#1}{#2}}{%
         \pgfmathparse{int(round(100*(#1/(0.274--0.355))-(-0.355*(100/(0.274--0.355)))))}%
        \xdef\tempa{\pgfmathresult}%
        \cellcolor{high!\tempa!low!\opacity} \hspace{-8pt}\ifdimcomp{#1pt}{>}{0pt}{\phantom{-}}{}\R{#1}{#2}\hspace{-8pt}
    }}
 }

\newcommand{\EnDeBLEU}[2]{%
    \ifdimcomp{#1pt}{>}{27.7 pt}{\R{#1}{#2}}{%
    \ifdimcomp{#1pt}{<}{19.1 pt}{\R{#1}{#2}}{%
         \pgfmathparse{int(round(100*(#1/(27.7-19.1))-(19.1*(100/(27.7-19.1)))))}%
        \xdef\tempa{\pgfmathresult}%
        \cellcolor{high!\tempa!low!\opacity} \hspace{-8pt}\ifdimcomp{#1pt}{>}{0pt}{\phantom{-}}{}\R{#1}{#2}\hspace{-8pt}
    }}
 }

\newcommand{\EnDechrF}[2]{%
    \ifdimcomp{#1pt}{>}{0.555 pt}{\R{#1}{#2}}{%
    \ifdimcomp{#1pt}{<}{0.498 pt}{\R{#1}{#2}}{%
         \pgfmathparse{int(round(100*(#1/(0.555-0.498))-(0.498*(100/(0.555-0.498)))))}%
        \xdef\tempa{\pgfmathresult}%
        \cellcolor{high!\tempa!low!\opacity} \hspace{-8pt}\ifdimcomp{#1pt}{>}{0pt}{\phantom{-}}{}\R{#1}{#2}\hspace{-8pt}
    }}
 }

\newcommand{\EnDeCOMET}[2]{%
    \ifdimcomp{#1pt}{>}{0.254 pt}{\R{#1}{#2}}{%
    \ifdimcomp{#1pt}{<}{-0.417 pt}{\R{#1}{#2}}{%
         \pgfmathparse{int(round(100*(#1/(0.254--0.417))-(-0.417*(100/(0.254--0.417)))))}%
        \xdef\tempa{\pgfmathresult}%
        \cellcolor{high!\tempa!low!\opacity} \hspace{-8pt}\ifdimcomp{#1pt}{>}{0pt}{\phantom{-}}{}\R{#1}{#2}\hspace{-8pt}
    }}
 }

\newcommand{\EnFrBLEU}[2]{%
    \ifdimcomp{#1pt}{>}{36.4 pt}{\R{#1}{#2}}{%
    \ifdimcomp{#1pt}{<}{27.4 pt}{\R{#1}{#2}}{%
         \pgfmathparse{int(round(100*(#1/(36.4-27.4))-(27.4*(100/(36.4-27.4)))))}%
        \xdef\tempa{\pgfmathresult}%
        \cellcolor{high!\tempa!low!\opacity} \hspace{-8pt}\ifdimcomp{#1pt}{>}{0pt}{\phantom{-}}{}\R{#1}{#2}\hspace{-8pt}
    }}
 }

\newcommand{\EnFrchrF}[2]{%
    \ifdimcomp{#1pt}{>}{0.619 pt}{\R{#1}{#2}}{%
    \ifdimcomp{#1pt}{<}{0.54 pt}{\R{#1}{#2}}{%
         \pgfmathparse{int(round(100*(#1/(0.619-0.54))-(0.54*(100/(0.619-0.54)))))}%
        \xdef\tempa{\pgfmathresult}%
        \cellcolor{high!\tempa!low!\opacity} \hspace{-8pt}\ifdimcomp{#1pt}{>}{0pt}{\phantom{-}}{}\R{#1}{#2}\hspace{-8pt}
    }}
 }

\newcommand{\EnFrCOMET}[2]{%
    \ifdimcomp{#1pt}{>}{0.408 pt}{\R{#1}{#2}}{%
    \ifdimcomp{#1pt}{<}{-0.039 pt}{\R{#1}{#2}}{%
         \pgfmathparse{int(round(100*(#1/(0.408--0.039))-(-0.039*(100/(0.408--0.039)))))}%
        \xdef\tempa{\pgfmathresult}%
        \cellcolor{high!\tempa!low!\opacity} \hspace{-8pt}\ifdimcomp{#1pt}{>}{0pt}{\phantom{-}}{}\R{#1}{#2}\hspace{-8pt}
    }}
 }

\newcommand{\ArEnBLEU}[2]{%
    \ifdimcomp{#1pt}{>}{29.7 pt}{\R{#1}{#2}}{%
    \ifdimcomp{#1pt}{<}{15.4 pt}{\R{#1}{#2}}{%
         \pgfmathparse{int(round(100*(#1/(29.7-15.4))-(15.4*(100/(29.7-15.4)))))}%
        \xdef\tempa{\pgfmathresult}%
        \cellcolor{high!\tempa!low!\opacity} \hspace{-8pt}\ifdimcomp{#1pt}{>}{0pt}{\phantom{-}}{}\R{#1}{#2}\hspace{-8pt}
    }}
 }

\newcommand{\ArEnchrF}[2]{%
    \ifdimcomp{#1pt}{>}{0.521 pt}{\R{#1}{#2}}{%
    \ifdimcomp{#1pt}{<}{0.413 pt}{\R{#1}{#2}}{%
         \pgfmathparse{int(round(100*(#1/(0.521-0.413))-(0.413*(100/(0.521-0.413)))))}%
        \xdef\tempa{\pgfmathresult}%
        \cellcolor{high!\tempa!low!\opacity} \hspace{-8pt}\ifdimcomp{#1pt}{>}{0pt}{\phantom{-}}{}\R{#1}{#2}\hspace{-8pt}
    }}
 }

\newcommand{\ArEnCOMET}[2]{%
    \ifdimcomp{#1pt}{>}{0.325 pt}{\R{#1}{#2}}{%
    \ifdimcomp{#1pt}{<}{-0.499 pt}{\R{#1}{#2}}{%
         \pgfmathparse{int(round(100*(#1/(0.325--0.499))-(-0.499*(100/(0.325--0.499)))))}%
        \xdef\tempa{\pgfmathresult}%
        \cellcolor{high!\tempa!low!\opacity} \hspace{-8pt}\ifdimcomp{#1pt}{>}{0pt}{\phantom{-}}{}\R{#1}{#2}\hspace{-8pt}
    }}
 }

\newcommand{\DeEnBLEU}[2]{%
    \ifdimcomp{#1pt}{>}{31.6 pt}{\R{#1}{#2}}{%
    \ifdimcomp{#1pt}{<}{23.2 pt}{\R{#1}{#2}}{%
         \pgfmathparse{int(round(100*(#1/(31.6-23.2))-(23.2*(100/(31.6-23.2)))))}%
        \xdef\tempa{\pgfmathresult}%
        \cellcolor{high!\tempa!low!\opacity} \hspace{-8pt}\ifdimcomp{#1pt}{>}{0pt}{\phantom{-}}{}\R{#1}{#2}\hspace{-8pt}
    }}
 }

\newcommand{\DeEnchrF}[2]{%
    \ifdimcomp{#1pt}{>}{0.554 pt}{\R{#1}{#2}}{%
    \ifdimcomp{#1pt}{<}{0.504 pt}{\R{#1}{#2}}{%
         \pgfmathparse{int(round(100*(#1/(0.554-0.504))-(0.504*(100/(0.554-0.504)))))}%
        \xdef\tempa{\pgfmathresult}%
        \cellcolor{high!\tempa!low!\opacity} \hspace{-8pt}\ifdimcomp{#1pt}{>}{0pt}{\phantom{-}}{}\R{#1}{#2}\hspace{-8pt}
    }}
 }

\newcommand{\DeEnCOMET}[2]{%
    \ifdimcomp{#1pt}{>}{0.379 pt}{\R{#1}{#2}}{%
    \ifdimcomp{#1pt}{<}{-0.082 pt}{\R{#1}{#2}}{%
         \pgfmathparse{int(round(100*(#1/(0.379--0.082))-(-0.082*(100/(0.379--0.082)))))}%
        \xdef\tempa{\pgfmathresult}%
        \cellcolor{high!\tempa!low!\opacity} \hspace{-8pt}\ifdimcomp{#1pt}{>}{0pt}{\phantom{-}}{}\R{#1}{#2}\hspace{-8pt}
    }}
 }

\newcommand{\FrEnBLEU}[2]{%
    \ifdimcomp{#1pt}{>}{36.2 pt}{\R{#1}{#2}}{%
    \ifdimcomp{#1pt}{<}{27.9 pt}{\R{#1}{#2}}{%
         \pgfmathparse{int(round(100*(#1/(36.2-27.9))-(27.9*(100/(36.2-27.9)))))}%
        \xdef\tempa{\pgfmathresult}%
        \cellcolor{high!\tempa!low!\opacity} \hspace{-8pt}\ifdimcomp{#1pt}{>}{0pt}{\phantom{-}}{}\R{#1}{#2}\hspace{-8pt}
    }}
 }

\newcommand{\FrEnchrF}[2]{%
    \ifdimcomp{#1pt}{>}{0.592 pt}{\R{#1}{#2}}{%
    \ifdimcomp{#1pt}{<}{0.54 pt}{\R{#1}{#2}}{%
         \pgfmathparse{int(round(100*(#1/(0.592-0.54))-(0.54*(100/(0.592-0.54)))))}%
        \xdef\tempa{\pgfmathresult}%
        \cellcolor{high!\tempa!low!\opacity} \hspace{-8pt}\ifdimcomp{#1pt}{>}{0pt}{\phantom{-}}{}\R{#1}{#2}\hspace{-8pt}
    }}
 }

\newcommand{\FrEnCOMET}[2]{%
    \ifdimcomp{#1pt}{>}{0.527 pt}{\R{#1}{#2}}{%
    \ifdimcomp{#1pt}{<}{0.078 pt}{\R{#1}{#2}}{%
         \pgfmathparse{int(round(100*(#1/(0.527-0.078))-(0.078*(100/(0.527-0.078)))))}%
        \xdef\tempa{\pgfmathresult}%
        \cellcolor{high!\tempa!low!\opacity} \hspace{-8pt}\ifdimcomp{#1pt}{>}{0pt}{\phantom{-}}{}\R{#1}{#2}\hspace{-8pt}
    }}
 }

\definecolor{high}{HTML}{ef3b2c}  
\definecolor{low}{HTML}{fff7bc}  
\newcommand*{\opacity}{70}

\newcommand{\R}[2]{\begin{minipage}{21pt}\centering\vspace{2pt} #1 \\ \tiny$\pm$#2\vspace{2pt}\end{minipage}}

\newcommand{\BLEU}{B\scalebox{0.5}{LEU}}
\newcommand{\chrF}{chrF}
\newcommand{\COMET}{C\scalebox{0.4}{OMET}}

\begin{adjustbox}{max width=\textwidth}
\begin{tabular}{llcc |ccc|ccc|ccc | ccc|ccc|ccc}
\toprule

\multirow{3}{*}{\rotatebox{90}{Model\hspace{10pt}}} & \multirow{2}{*}{Enc.} &\multirow{2}{*}{Dec.} & \multirow{3}{*}{\begin{minipage}{30pt}\centering\vspace{5pt} Char. proc. params\end{minipage}} &
\multicolumn{9}{c}{From English} & \multicolumn{9}{c}{Into English} \\  \cmidrule(lr){5-13}  \cmidrule(lr){14-22}
& & &
& \multicolumn{3}{c}{ar}
& \multicolumn{3}{c}{de}
& \multicolumn{3}{c}{fr}
& \multicolumn{3}{c}{ar}
& \multicolumn{3}{c}{de}
& \multicolumn{3}{c}{fr} \\
\cmidrule(lr){2-3} \cmidrule(lr){5-7} \cmidrule(lr){8-10} \cmidrule(lr){11-13} \cmidrule(lr){14-16} \cmidrule(lr){17-19} \cmidrule(lr){20-22}

& \multicolumn{2}{c}{downsample} &
& \BLEU & \chrF & \COMET
& \BLEU & \chrF & \COMET
& \BLEU & \chrF & \COMET
& \BLEU & \chrF & \COMET
& \BLEU & \chrF & \COMET
& \BLEU & \chrF & \COMET \\ \midrule

\multicolumn{3}{l}{BPE 16k} & 16516
& \EnArBLEU{11.2}{0.2}
& \EnArchrF{.436}{.002}
& \EnArCOMET{.258}{.011}
& \EnDeBLEU{27.7}{0.3}
& \EnDechrF{.555}{.002}
& \EnDeCOMET{.254}{.005}
& \EnFrBLEU{36.4}{0.3}
& \EnFrchrF{.619}{.002}
& \EnFrCOMET{.408}{.008}
& \ArEnBLEU{29.7}{0.2}
& \ArEnchrF{.521}{.001}
& \ArEnCOMET{.325}{.147}
& \DeEnBLEU{31.6}{0.3}
& \DeEnchrF{.554}{.001}
& \DeEnCOMET{.379}{.008}
& \FrEnBLEU{36.2}{0.3}
& \FrEnchrF{.592}{.003}
& \FrEnCOMET{.527}{.005}
\\ 
\multicolumn{3}{l}{Vanilla char.} & 658
& \EnArBLEU{13.5}{0.4}
& \EnArchrF{.447}{.004}
& \EnArCOMET{.267}{.016}
& \EnDeBLEU{25.6}{0.7}
& \EnDechrF{.550}{.005}
& \EnDeCOMET{.165}{.034}
& \EnFrBLEU{34.6}{0.7}
& \EnFrchrF{.611}{.002}
& \EnFrCOMET{.350}{.020}
& \ArEnBLEU{27.7}{0.8}
& \ArEnchrF{.518}{.006}
& \ArEnCOMET{.238}{.034}
& \DeEnBLEU{29.4}{0.7}
& \DeEnchrF{.545}{.005}
& \DeEnCOMET{.327}{.029}
& \FrEnBLEU{34.7}{0.4}
& \FrEnchrF{.585}{.003}
& \FrEnCOMET{.487}{.012}
\\ \midrule
\multirow{6}{*}{\rotatebox{90}{Lee-style}} & 3 & --- & 9672
& \EnArBLEU{13.1}{0.5}
& \EnArchrF{.448}{.002}
& \EnArCOMET{.274}{.009}
& \EnDeBLEU{25.9}{0.7}
& \EnDechrF{.552}{.001}
& \EnDeCOMET{.200}{.023}
& \EnFrBLEU{35.2}{0.4}
& \EnFrchrF{.613}{.002}
& \EnFrCOMET{.383}{.010}
& \ArEnBLEU{28.0}{0.4}
& \ArEnchrF{.521}{.002}
& \ArEnCOMET{.257}{.015}
& \DeEnBLEU{30.2}{0.5}
& \DeEnchrF{.551}{.003}
& \DeEnCOMET{.345}{.022}
& \FrEnBLEU{35.3}{0.2}
& \FrEnchrF{.588}{.001}
& \FrEnCOMET{.506}{.013}
\\ 
 & 5 & --- & 9672
& \EnArBLEU{12.5}{0.1}
& \EnArchrF{.439}{.002}
& \EnArCOMET{.245}{.013}
& \EnDeBLEU{25.0}{0.4}
& \EnDechrF{.545}{.002}
& \EnDeCOMET{.140}{.013}
& \EnFrBLEU{33.2}{0.1}
& \EnFrchrF{.602}{.003}
& \EnFrCOMET{.303}{.017}
& \ArEnBLEU{24.9}{4.4}
& \ArEnchrF{.491}{.042}
& \ArEnCOMET{.090}{.228}
& \DeEnBLEU{28.9}{0.3}
& \DeEnchrF{.543}{.002}
& \DeEnCOMET{.311}{.019}
& \FrEnBLEU{34.4}{0.3}
& \FrEnchrF{.583}{.002}
& \FrEnCOMET{.483}{.016}
\\ 
 & 3 & 3 & 9646
& \EnArBLEU{11.0}{0.2}
& \EnArchrF{.432}{.002}
& \EnArCOMET{.143}{.013}
& \EnDeBLEU{23.4}{0.4}
& \EnDechrF{.541}{.002}
& \EnDeCOMET{.065}{.028}
& \EnFrBLEU{31.7}{0.5}
& \EnFrchrF{.603}{.002}
& \EnFrCOMET{.277}{.012}
& \ArEnBLEU{25.6}{0.3}
& \ArEnchrF{.509}{.001}
& \ArEnCOMET{.170}{.016}
& \DeEnBLEU{28.0}{0.3}
& \DeEnchrF{.537}{.002}
& \DeEnCOMET{.262}{.019}
& \FrEnBLEU{33.3}{0.4}
& \FrEnchrF{.577}{.001}
& \FrEnCOMET{.440}{.015}
\\ 
 & 5 & 5 & 9646
& \EnArBLEU{9.4}{0.5}
& \EnArchrF{.418}{.003}
& \EnArCOMET{.006}{.015}
& \EnDeBLEU{21.8}{0.3}
& \EnDechrF{.524}{.002}
& \EnDeCOMET{-.106}{.021}
& \EnFrBLEU{28.7}{1.7}
& \EnFrchrF{.584}{.011}
& \EnFrCOMET{.094}{.096}
& \ArEnBLEU{23.7}{0.3}
& \ArEnchrF{.492}{.001}
& \ArEnCOMET{.033}{.015}
& \DeEnBLEU{25.5}{0.3}
& \DeEnchrF{.519}{.003}
& \DeEnCOMET{.131}{.019}
& \FrEnBLEU{30.9}{0.5}
& \FrEnchrF{.561}{.004}
& \FrEnCOMET{.335}{.018}
\\ \midrule
\multirow{6}{*}{\rotatebox{90}{Charformer}} & 3 & --- & 1320
& \EnArBLEU{13.3}{0.3}
& \EnArchrF{.448}{.002}
& \EnArCOMET{.261}{.011}
& \EnDeBLEU{25.9}{0.5}
& \EnDechrF{.550}{.004}
& \EnDeCOMET{.167}{.026}
& \EnFrBLEU{32.9}{0.3}
& \EnFrchrF{.607}{.003}
& \EnFrCOMET{.300}{.018}
& \ArEnBLEU{27.3}{0.5}
& \ArEnchrF{.520}{.002}
& \ArEnCOMET{.229}{.028}
& \DeEnBLEU{29.9}{0.3}
& \DeEnchrF{.548}{.001}
& \DeEnCOMET{.327}{.008}
& \FrEnBLEU{35.1}{0.3}
& \FrEnchrF{.588}{.002}
& \FrEnCOMET{.495}{.013}
\\ 
 & 5 & --- & 1320
& \EnArBLEU{12.2}{0.3}
& \EnArchrF{.435}{.002}
& \EnArCOMET{.179}{.020}
& \EnDeBLEU{24.2}{0.6}
& \EnDechrF{.535}{.003}
& \EnDeCOMET{.060}{.027}
& \EnFrBLEU{31.3}{0.4}
& \EnFrchrF{.591}{.003}
& \EnFrCOMET{.171}{.026}
& \ArEnBLEU{25.1}{0.6}
& \ArEnchrF{.500}{.002}
& \ArEnCOMET{.103}{.022}
& \DeEnBLEU{28.1}{0.4}
& \DeEnchrF{.535}{.003}
& \DeEnCOMET{.227}{.022}
& \FrEnBLEU{33.7}{0.2}
& \FrEnchrF{.577}{.002}
& \FrEnCOMET{.428}{.012}
\\ 
 & 3 & 3 & 1165
& \EnArBLEU{10.3}{0.5}
& \EnArchrF{.431}{.004}
& \EnArCOMET{.000}{.000}
& \EnDeBLEU{23.2}{0.5}
& \EnDechrF{.540}{.004}
& \EnDeCOMET{.037}{.034}
& \EnFrBLEU{30.6}{0.4}
& \EnFrchrF{.601}{.003}
& \EnFrCOMET{.192}{.031}
& \ArEnBLEU{24.5}{0.4}
& \ArEnchrF{.506}{.003}
& \ArEnCOMET{.125}{.021}
& \DeEnBLEU{27.5}{0.5}
& \DeEnchrF{.538}{.003}
& \DeEnCOMET{.225}{.021}
& \FrEnBLEU{32.6}{0.3}
& \FrEnchrF{.576}{.001}
& \FrEnCOMET{.425}{.014}
\\ 
 & 5 & 5 & 1165
& \EnArBLEU{8.4}{0.2}
& \EnArchrF{.402}{.003}
& \EnArCOMET{-.121}{.023}
& \EnDeBLEU{19.9}{0.2}
& \EnDechrF{.510}{.002}
& \EnDeCOMET{-.250}{.027}
& \EnFrBLEU{27.4}{0.7}
& \EnFrchrF{.575}{.005}
& \EnFrCOMET{-.039}{.029}
& \ArEnBLEU{18.4}{3.1}
& \ArEnchrF{.448}{.029}
& \ArEnCOMET{-.248}{.173}
& \DeEnBLEU{23.5}{0.5}
& \DeEnchrF{.511}{.003}
& \DeEnCOMET{.018}{.029}
& \FrEnBLEU{29.2}{0.7}
& \FrEnchrF{.552}{.002}
& \FrEnCOMET{.228}{.035}
\\ \midrule
\multirow{6}{*}{\rotatebox{90}{Canine}} & 3 & --- & 6446
& \EnArBLEU{12.6}{0.3}
& \EnArchrF{.440}{.002}
& \EnArCOMET{.195}{.019}
& \EnDeBLEU{25.4}{0.5}
& \EnDechrF{.547}{.002}
& \EnDeCOMET{.121}{.024}
& \EnFrBLEU{33.2}{0.6}
& \EnFrchrF{.606}{.004}
& \EnFrCOMET{.269}{.024}
& \ArEnBLEU{26.1}{0.5}
& \ArEnchrF{.512}{.004}
& \ArEnCOMET{.137}{.024}
& \DeEnBLEU{29.1}{0.4}
& \DeEnchrF{.546}{.002}
& \DeEnCOMET{.273}{.020}
& \FrEnBLEU{34.5}{0.4}
& \FrEnchrF{.583}{.003}
& \FrEnCOMET{.448}{.014}
\\ 
 & 5 & --- & 7470
& \EnArBLEU{11.2}{0.2}
& \EnArchrF{.421}{.001}
& \EnArCOMET{.045}{.005}
& \EnDeBLEU{22.5}{0.4}
& \EnDechrF{.524}{.004}
& \EnDeCOMET{-.095}{.027}
& \EnFrBLEU{30.5}{0.5}
& \EnFrchrF{.584}{.004}
& \EnFrCOMET{.273}{.029}
& \ArEnBLEU{22.1}{0.6}
& \ArEnchrF{.477}{.001}
& \ArEnCOMET{-.121}{.023}
& \DeEnBLEU{27.3}{0.3}
& \DeEnchrF{.528}{.001}
& \DeEnCOMET{.115}{.022}
& \FrEnBLEU{32.5}{0.5}
& \FrEnchrF{.566}{.004}
& \FrEnCOMET{.273}{.029}
\\ 
 & 3 & 3 & 6291
& \EnArBLEU{10.3}{0.5}
& \EnArchrF{.425}{.004}
& \EnArCOMET{.064}{.023}
& \EnDeBLEU{22.4}{0.3}
& \EnDechrF{.534}{.003}
& \EnDeCOMET{-.034}{.023}
& \EnFrBLEU{30.2}{0.5}
& \EnFrchrF{.595}{.004}
& \EnFrCOMET{.139}{.027}
& \ArEnBLEU{23.7}{0.9}
& \ArEnchrF{.499}{.006}
& \ArEnCOMET{.034}{.030}
& \DeEnBLEU{25.9}{1.0}
& \DeEnchrF{.527}{.008}
& \DeEnCOMET{.127}{.043}
& \FrEnBLEU{32.5}{0.3}
& \FrEnchrF{.575}{.002}
& \FrEnCOMET{.368}{.008}
\\ 
 & 5 & 5 & 7444
& \EnArBLEU{6.9}{0.4}
& \EnArchrF{.373}{.007}
& \EnArCOMET{-.355}{.041}
& \EnDeBLEU{19.1}{0.4}
& \EnDechrF{.498}{.005}
& \EnDeCOMET{-.417}{.036}
& \EnFrBLEU{27.9}{0.6}
& \EnFrchrF{.540}{.001}
& \EnFrCOMET{.078}{.019}
& \ArEnBLEU{15.4}{0.3}
& \ArEnchrF{.413}{.003}
& \ArEnCOMET{-.499}{.030}
& \DeEnBLEU{23.2}{0.2}
& \DeEnchrF{.504}{.001}
& \DeEnCOMET{-.082}{.010}
& \FrEnBLEU{27.9}{0.6}
& \FrEnchrF{.540}{.001}
& \FrEnCOMET{.078}{.019}
\\ 
\bottomrule
\end{tabular}
\end{adjustbox}

	\caption{Translation quality of the models on the IWSLT data. The fourth
		column shows the size of the character-processing layers expressed as
		the vocabulary size of Transformer Base having the same number of
		parameters in the embeddings.}\label{tab:iwsltData}

\end{table*}

We implement the models using Huggingface Transformers
\citep{wolf-etal-2020-transformers}. We take the CANINE layer from Huggingface
Transformers and use an independent implementation of
Charformer\footnote{\url{https://github.com/lucidrains/charformer-pytorch}}.
Our source
code is available on Github.\footnote{\url{https://github.com/jlibovicky/char-nmt-two-step-decoder}}
Hyperparameters and other experimental details 
can be found 
in Appendix~\ref{appendix:iwslt}.

\subsection{Experimental Setup}

We evaluate the models on translation between English paired with German,
French, and Arabic (with English as both input and output) using the IWSLT 2017 datasets
\citep{cettolo2017overview} with a training data size of around 200k sentences
for each language pair (see Appendix~\ref{appendix:iwslt} for details).

For the subword models, we tokenize the input using the Moses tokenizer
\citep{koehn-etal-2007-moses} and then further split the words into subword
units using BPE \citep{sennrich-etal-2016-neural} with 16k merge operations.
For the character models, we limit the vocabulary to 300 UTF-8 characters.

We use the Transformer Base architecture \citep{vaswani2017attention} in all
experiments. We
make no changes to it 
in the subword and baseline character
experiments. 
In the later experiments, we 
replace the embedding lookup with the character processing architectures.
For the Lee-style encoder, we chose
similar hyperparameters as related work \citep{banar2020character}.
For experiments with Charformer and CANINE models, we set the hyperparameters
such that they cover the same character span before downsampling as the
Lee-style encoder, which causes the models to have fewer parameters than a
Lee-style encoder. Note however that for both the Charformer and the CANINE
models, the number of parameters is almost independent of the character window
width. For all three character processing architectures, we experiment with
downsampling factors of 3 and 5 (a 16k BPE vocabulary corresponds to a
downsampling factor of about 4 in English).

\subsection{Translation Quality}

We evaluate the translation quality using the BLEU score
\citep{papineni-etal-2002-bleu}, the chrF score \citep{popovic-2015-chrf} (as
implemented in SacreBLEU\@; \citealp{post-2018-call}),\footnote{BLEU score signature {\tt nrefs:1|\hskip0pt
case:mixed|\hskip0pt eff:no|\hskip0pt tok:13a|\hskip0pt smooth:exp|\hskip0pt
version:2.0.0} chrF score signature {\tt nrefs:1|\hskip0pt case:mixed|\hskip0pt
eff:yes|\hskip0pt nc:6|\hskip0pt nw:0|\hskip0pt space:no|\hskip0pt
version:2.0.0}} and the COMET score
\citep{rei-etal-2020-comet}. We run each experiment 4 times and report the mean
value and standard deviation.

The results are presented in Table~\ref{tab:iwsltData}. Except for translation
into Arabic, where
character methods outperform BPEs
 (which is consistent with the findings of \citealp{shaham-levy-2021-neural} and \citealp{li-etal-2021-char}),
subword methods are always better than characters.

%
The Lee-style encoder outperforms the two more recent methods and the method of
using the character embeddings directly. Charformer performs similarly to using
character embeddings directly, CANINE is significantly worse. The results are
mostly consistent across the language pairs.

Increasing the downsampling 
rate 
from 3 to 5 degrades the translation quality for
all architectures. Employing the two-step decoder matches the decoding speed
of subword models. However, the overall translation quality is much worse.

The three metrics that we use give consistent results in most cases. Often,
relatively small differences in BLEU and chrF scores correspond to much bigger
differences in the COMET score.

\subsection{Inference}

\enote{AF}{It would be great to introduce this section with one or two sentences summarizing why it is important and how it is related to the main theme of the paper; you could also preview the final result of the subsection already here.}
\enote{AF}{The ordering here is maybe not optimal, this should maybe be presented after the big scale translation quality experiment. It is probably too late to think about this, but I suspect that this subsection could actually mostly go into the appendix.}

Inference algorithms for neural MT have been discussed extensively
\citep{meister-etal-2020-beam,massarelli-etal-2020-decoding,shi2020why,shaham2021what}
for the subword models. Subword translation quality quickly degrades beyond a
certain beam width unless heuristically defined length normalization is
applied.

As an alternative, \citet{eikema-aziz-2020-map} recently proposed Minimum Bayes
Risk (MBR\@; \citealt{goel2000minimum}) estimation as an alternative. Assuming
that similar sentences should be similarly probable, they propose repeatedly
sampling from the model and selecting a sentence that is most similar to other
samples. With subword models, MBR performs comparably to beam search.





Intuitive arguments about the inference algorithms are often based on the
properties of the subword output distribution. On average, character models
will produce distributions with lower perplexity and thus likely suffer more
from the exposure bias which might harm sampling from the model.
Therefore, there is a risk that these empirical findings do not apply to
character-level models.

We explore what decoding strategies are best suited for the character-level
models. We compare the translation quality of beam search decoding with
different degrees of length normalization.\footnote{As we increase beam size,
the number of search errors is decreasing, but here we are evaluating modeling
errors, not search errors.} Further, we compare length-normalized beam search
decoding with MBR (with 100 samples), greedy decoding, and random sampling. We
use the chrF as a comparison metric which allows pre-computing the character
$n$-grams and thus faster sentence pair comparison than the originally proposed
METEOR \citep{denkowski-lavie-2011-meteor}.

\begin{figure}
    \centering
    \begin{minipage}{.65\columnwidth}
        \includegraphics{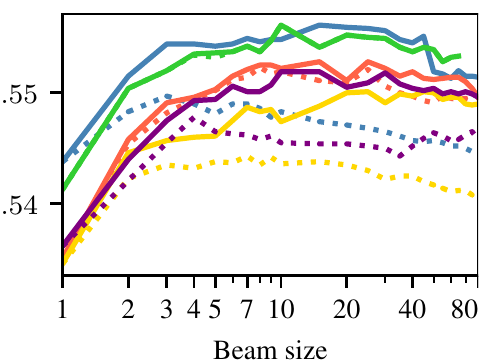}
    \end{minipage}\begin{minipage}{.34\columnwidth}
        \footnotesize
            {\scriptsize\color[HTML]{4682B4}$\blacksquare$} BPE          \hfill -0.94 \\
            {\scriptsize\color[HTML]{32CD32}$\blacksquare$} Char direct  \hfill  0.04 \\
            {\scriptsize\color[HTML]{FF6347}$\blacksquare$} Lee-style    \hfill  0.00 \\
            {\scriptsize\color[HTML]{FFD700}$\blacksquare$} Charformer   \hfill -0.91 \\
            {\scriptsize\color[HTML]{800080}$\blacksquare$} CANINE       \hfill  0.15 \\

            \vspace{1cm}

    \end{minipage}


    \caption{chrF scores for IWSLT en-de translation for different models and
    beam sizes. The dotted lines are without length normalization, the solid
    lines are with length normalization. All character processing architectures
    use a downsampling window of size 3. The legend tabulates the Pearson
    correlation of the beam size (starting from 5) and the chrF
    score.}\label{fig:beamSize}

\end{figure}

Figure~\ref{fig:beamSize} shows the translation quality of the selected models
for different beam sizes. The dotted lines denoting the translation quality
without length normalization show that the quality of the subword models
quickly deteriorates without length normalization, whereas vanilla and
Lee-style character-level models do not seem to suffer from this problem.

%

\begin{table}[!t]

    \centering
\newcommand{\chrfVal}[1]{%
    \ifdimcomp{#1pt}{>}{0.555 pt}{#1}{%
    \ifdimcomp{#1pt}{<}{0.215 pt}{#1}{%
         \pgfmathparse{int(round(100*(#1/(0.555-0.215))-(0.215*(100/(0.555-0.215)))))}%
        \xdef\tempa{\pgfmathresult}%
        \cellcolor{highGreen!\tempa!low!\opacity} \hspace{-8pt}\ifdimcomp{#1pt}{>}{0pt}{\phantom{-}}{}#1\hspace{-8pt}
    }}
 }

\newcommand{\cometVal}[1]{%
    \ifdimcomp{#1pt}{>}{0.262 pt}{#1}{%
    \ifdimcomp{#1pt}{<}{-1.79 pt}{#1}{%
         \pgfmathparse{int(round(100*(#1/(0.262--1.79))-(-1.79*(100/(0.262--1.79)))))}%
        \xdef\tempa{\pgfmathresult}%
        \cellcolor{highRed!\tempa!low!\opacity} \hspace{-8pt}\ifdimcomp{#1pt}{>}{0pt}{\phantom{-}}{}#1\hspace{-8pt}
    }}
 }

\newcommand{\minval}{-1.79}
\newcommand{\maxval}{0.262}
\definecolor{highRed}{HTML}{ef3b2c}  
\definecolor{highGreen}{HTML}{6ba35b}  
\definecolor{low}{HTML}{fff7bc}  
\newcommand*{\opacity}{70}

\newcommand{\Dec}[1]{%
    \ifdimcomp{#1pt}{>}{\maxval pt}{#1}{%
    \ifdimcomp{#1pt}{<}{\minval pt}{#1}{%
        \pgfmathparse{int(round(100*(#1/(\maxval-\minval))-(\minval*(100/(\maxval-\minval)))))}%
        \xdef\tempa{\pgfmathresult}%
        \cellcolor{high!\tempa!low!\opacity} \hspace{1pt}\ifdimcomp{#1pt}{>}{0pt}{\phantom{-}}{}#1\hspace{1pt}
    }}
 }


\newcommand{\ColumnTit}[1]{\multicolumn{1}{p{20pt}}{\centering #1}}

\begin{adjustbox}{max width=\columnwidth}
\footnotesize
\begin{tabular}{lcc cccc}
		\toprule
\multirow{2}{*}{\rotatebox{90}{Model}}
		& Enc. & Dec. & \ColumnTit{Sample} & \ColumnTit{Greedy} & \ColumnTit{Beam} & \ColumnTit{MBR} \\ \cmidrule(lr){2-3}

		& \multicolumn{2}{c}{downsample} \\
\midrule

\multicolumn{3}{l}{BPE 16k}
& \chrfVal{0.482}
& \chrfVal{0.545}
& \chrfVal{0.555}
& \chrfVal{0.554}
\\
& &
& \cometVal{-0.132}
& \cometVal{0.199}
& \cometVal{0.262}
& \cometVal{0.187}
\\ \midrule
\multicolumn{3}{l}{Vanilla char.}
& \chrfVal{0.448}
& \chrfVal{0.537}
& \chrfVal{0.537}
& \chrfVal{0.538}
\\
& &
& \cometVal{-0.446}
& \cometVal{0.117}
& \cometVal{0.165}
& \cometVal{0.086}
\\ \midrule
\multirow{8}{*}{\rotatebox{90}{Lee-style}} & \multirow{2}{*}{3} & \multirow{2}{*}{---}
& \chrfVal{0.461}
& \chrfVal{0.539}
& \chrfVal{0.552}
& \chrfVal{0.544}
\\
& &
& \cometVal{-0.340}
& \cometVal{0.142}
& \cometVal{0.200}
& \cometVal{0.106}
\\ \cmidrule{2-7}
 & \multirow{2}{*}{5} & \multirow{2}{*}{---}
& \chrfVal{0.454}
& \chrfVal{0.527}
& \chrfVal{0.545}
& \chrfVal{0.537}
\\
& &
& \cometVal{-0.371}
& \cometVal{0.082}
& \cometVal{0.140}
& \cometVal{0.051}
\\ \cmidrule{2-7}
 & \multirow{2}{*}{3} & \multirow{2}{*}{3}
& \chrfVal{0.430}
& \chrfVal{0.523}
& \chrfVal{0.540}
& \chrfVal{0.526}
\\
& &
& \cometVal{-0.657}
& \cometVal{-0.015}
& \cometVal{0.065}
& \cometVal{-0.105}
\\ \cmidrule{2-7}
 & \multirow{2}{*}{5} & \multirow{2}{*}{5}
& \chrfVal{0.395}
& \chrfVal{0.495}
& \chrfVal{0.524}
& \chrfVal{0.500}
\\
& &
& \cometVal{-0.918}
& \cometVal{-0.259}
& \cometVal{-0.106}
& \cometVal{-0.341}
\\ \midrule
\multirow{8}{*}{\rotatebox{90}{Charformer}} & \multirow{2}{*}{3} & \multirow{2}{*}{---}
& \chrfVal{0.305}
& \chrfVal{0.530}
& \chrfVal{0.547}
& \chrfVal{0.448}
\\
& &
& \cometVal{-1.490}
& \cometVal{0.061}
& \cometVal{0.149}
& \cometVal{-0.831}
\\ \cmidrule{2-7}
 & \multirow{2}{*}{5} & \multirow{2}{*}{---}
& \chrfVal{0.291}
& \chrfVal{0.515}
& \chrfVal{0.536}
& \chrfVal{0.442}
\\
& &
& \cometVal{-1.480}
& \cometVal{-0.017}
& \cometVal{0.069}
& \cometVal{-0.854}
\\ \cmidrule{2-7}
 & \multirow{2}{*}{3} & \multirow{2}{*}{3}
& \chrfVal{0.227}
& \chrfVal{0.462}
& \chrfVal{0.540}
& \chrfVal{0.412}
\\
& &
& \cometVal{-1.720}
& \cometVal{-0.424}
& \cometVal{0.036}
& \cometVal{-1.090}
\\ \cmidrule{2-7}
 & \multirow{2}{*}{5} & \multirow{2}{*}{5}
& \chrfVal{0.215}
& \chrfVal{0.478}
& \chrfVal{0.510}
& \chrfVal{0.371}
\\
& &
& \cometVal{-1.790}
& \cometVal{-0.733}
& \cometVal{-0.230}
& \cometVal{-1.290}
\\ \midrule
\multirow{8}{*}{\rotatebox{90}{Canine}} & \multirow{2}{*}{3} & \multirow{2}{*}{---}
& \chrfVal{0.307}
& \chrfVal{0.531}
& \chrfVal{0.547}
& \chrfVal{0.456}
\\
& &
& \cometVal{-1.500}
& \cometVal{0.051}
& \cometVal{0.121}
& \cometVal{-0.838}
\\ \cmidrule{2-7}
 & \multirow{2}{*}{5} & \multirow{2}{*}{---}
& \chrfVal{0.301}
& \chrfVal{0.509}
& \chrfVal{0.524}
& \chrfVal{0.440}
\\
& &
& \cometVal{-1.530}
& \cometVal{-0.163}
& \cometVal{-0.095}
& \cometVal{-0.968}
\\ \cmidrule{2-7}
 & \multirow{2}{*}{3} & \multirow{2}{*}{3}
& \chrfVal{0.253}
& \chrfVal{0.516}
& \chrfVal{0.534}
& \chrfVal{0.413}
\\
& &
& \cometVal{-1.680}
& \cometVal{-0.097}
& \cometVal{-0.034}
& \cometVal{-1.130}
\\ \cmidrule{2-7}
 & \multirow{2}{*}{5} & \multirow{2}{*}{5}
& \chrfVal{0.219}
& \chrfVal{0.475}
& \chrfVal{0.498}
& \chrfVal{0.371}
\\
& &
& \cometVal{-1.750}
& \cometVal{-0.524}
& \cometVal{-0.417}
& \cometVal{-1.370}
\\ 
\bottomrule
\end{tabular}
\end{adjustbox}

		\caption{chrF (yellow-green scale) and COMET (yellow-red scale)
		scores for decoding methods for models trained on en-de
		systems.}\label{tab:decoding}

\end{table}

Table~\ref{tab:decoding} presents the translation quality for different
decoding methods. In all cases, beam search is the best strategy. Sampling from
character-level models leads to very poor translation quality that in turn also
influences the MBR decoding 
leading
to much worse results than beam search.

Our experiments show that beam search with length normalization is the best
inference algorithm for character-level models. They also seem to be more
resilient towards the beam search curse compared to subword models.

%
%

%

\section{Experiments on WMT Data}

\begin{table*}

    \centering\newcommand{\BLEU}{B\scalebox{0.5}{LEU}}
\newcommand{\chrF}{chrF}
\newcommand{\COMET}{C\scalebox{0.4}{OMET}}

\newcommand{\R}[2]{\begin{minipage}{16pt}\centering #1 \\ \tiny \scalebox{.6}{$\pm$}#2\vspace*{2pt}\end{minipage}}
\newcommand{\T}[2]{\begin{minipage}{16pt}\centering #1 \\ \tiny vs.~#2\vspace*{2pt}\end{minipage}}

\begin{adjustbox}{max width=\textwidth}
\footnotesize
\begin{tabular}{ll ccc ccc ccc c c cc c}

\toprule

    &
    & \multicolumn{3}{c}{News}
    & \multicolumn{3}{c}{IT}
    & \multicolumn{3}{c}{Medical}
    & \multirow{2}{*}{\begin{minipage}{1cm}\centering\vspace{1pt} Gender \\ Acc.\end{minipage}}
    & \multirow{2}{*}{\begin{minipage}{1cm}\centering\vspace{-1pt} Avg. Morpheval\end{minipage}}
    & \multicolumn{2}{c}{Recall of novel}
    & \multirow{2}{*}{\begin{minipage}{1cm}\centering\vspace{-1pt} Noisy\\set\\\chrF\end{minipage}}\\

\cmidrule(lr){3-5}
\cmidrule(lr){6-8}
\cmidrule(lr){9-11}
\cmidrule(lr){14-15}

    &
    & \BLEU & \chrF & \COMET
    & \BLEU & \chrF & \COMET
    & \BLEU & \chrF & \COMET
    & &
    & F\scalebox{0.5}{orms} & L\scalebox{0.5}{emmas} \\

\midrule

\multirow{4}{*}{\rotatebox{90}{en-cs\phantom{d}~~~}}
    & BPE 16k &
    \R{30.8}{0.8} & \R{.585}{.006} & \R{.672}{.022} & \R{34.5}{1.3} & \R{.623}{.008} & \R{.889}{.022} & \R{26.4}{1.4} & \R{.519}{.010} & \R{.734}{.037} & 71.3 & 86.6 & \T{33.7}{63.7} & \T{48.5}{71.1} & \R{.436}{.002}
\\
    & BPE to char. &
    \R{28.4}{0.8} & \R{.570}{.006} & \R{.597}{.024} & \R{31.4}{1.2} & \R{.603}{.008} & \R{.821}{.025} & \R{23.6}{1.3} & \R{.499}{.010} & \R{.674}{.039} & 68.9 & 87.0 & \T{34.3}{} & \T{47.4}{} & \R{.436}{.001}
\\
    & Vanilla char. &
    \R{27.7}{0.7} & \R{.563}{.006} & \R{.550}{.026} & \R{30.0}{1.2} & \R{.589}{.008} & \R{.778}{.028} & \R{23.3}{1.3} & \R{.492}{.010} & \R{.663}{.039} & 70.2 & 86.4 & \T{34.4}{61.0} & \T{47.4}{68.7} & \R{.493}{.001}
\\
    & Lee-style enc. &
    \R{28.8}{0.8} & \R{.568}{.006} & \R{.609}{.024} & \R{31.7}{1.3} & \R{.606}{.008} & \R{.849}{.024} & \R{24.3}{1.3} & \R{.506}{.010} & \R{.696}{.038} & 65.6 & 86.6 & \T{34.1}{61.7} & \T{48.5}{69.2} & \R{.497}{.001}
\\

\midrule

\multirow{4}{*}{\rotatebox{90}{en-de~~~}}
    & BPE 16k &
    \R{31.5}{0.9} & \R{.603}{.006} & \R{.418}{.021} & \R{45.6}{1.3} & \R{.701}{.009} & \R{.622}{.021} & \R{38.7}{1.6} & \R{.640}{.010} & \R{.569}{.034} & 66.5 & 90.6 & \T{40.2}{72.3} & \T{51.0}{67.0} & \R{.464}{.002}
\\
    & BPE to char. &
    \R{29.1}{0.8} & \R{.589}{.006} & \R{.360}{.022} & \R{46.5}{1.3} & \R{.703}{.008} & \R{.617}{.021} & \R{36.0}{1.4} & \R{.621}{.009} & \R{.513}{.035} & 71.2 & 91.3 & \T{45.1}{71.1} & \T{50.8}{65.5} & .465
    \\

    & Vanilla char. &
    \R{27.8}{0.8} & \R{.578}{.006} & \R{.321}{.023} & \R{45.3}{1.3} & \R{.698}{.008} & \R{.600}{.022} & \R{35.6}{1.4} & \R{.618}{.009} & \R{.496}{.036} & 71.2 & 91.4 & \T{50.7}{64.3} & \T{45.1}{70.2} & \R{.504}{.001}
\\
    & Lee-style enc. &
    \R{29.1}{0.8} & \R{.588}{.006} & \R{.363}{.022} & \R{46.5}{1.3} & \R{.710}{.008} & \R{.619}{.022} & \R{36.5}{1.4} & \R{.623}{.009} & \R{.500}{.037} & 74.0 & 91.5 & \T{44.5}{77.1} & \T{50.8}{65.5} & \R{.515}{.001}
\\

\bottomrule

\end{tabular}
\end{adjustbox}

    \caption{Results of the WMT-scale experiments.}\label{tab:wmtResults}

\end{table*}

Based on the results of the experiments with the IWSLT data, we further
experiment only with the Lee-style encoder using a downsampling factor of 3
on the source side. Additionally, we experiment with hybrid systems with a
subword encoder and character decoder. We train translation systems of
competitive quality on two high-resource language pairs, English-Czech and
English-German, and perform an extensive evaluation.

\subsection{Experimental Setup}

For English-to-Czech translation, we use the CzEng 2.0 corpus
\citep{kocmi2020czeng} that aggregates and curates all sources for this
language pair. We use all 66M authentic parallel sentence pairs and 50M
back-translated Czech sentences.

For the English-to-German translation, we use a subset of the training data
used by \citet{chen-etal-2021-university}. The data consists of 66M authentic
sentence pairs filtered from the available data for WMT and 52M back-translated
German sentences from News Crawl 2020.

We tag the back-translation data \citep{caswell-etal-2019-tagged}. We use the
Transformer Big architecture for all experiments with hyperparameters following
\citet{popel2018training}. For the Lee-style encoder, we double the hidden
layer sizes compared to the IWSLT experiments (following the hidden size
increase between the Transformer Base and Big architectures).
In contrast to the previous set of experiments, we use Fairseq
\citep{ott-etal-2019-fairseq}. Our code is available on
Github\footnote{\url{https://github.com/jlibovicky/char-nmt-fairseq}}.
System outputs are attached to the paper in the ACL anthology.

We evaluate the systems not only on WMT20 test sets but also on data that often
motivated the research of character-level methods.
We evaluate the out-of-domain performance of the models on the NHS test set
from the WMT17 Biomedical Task \citep{jimeno-yepes-etal-2017-findings} and on
the WMT16 IT Domain test set \citep{bojar-etal-2016-findings}. We use the same
evaluation metrics as for the IWSLT experiments. We estimate the confidence
intervals using bootstrap resampling \citep{koehn-2004-statistical}.

We also assess the gender bias of the systems
\citep{stanovsky-etal-2019-evaluating,kocmi-etal-2020-gender}, using a dataset
of sentence pairs with stereotypical and non-stereotypical English sentences.
We measure the accuracy of gendered nouns and pronouns using word alignment and
morphological analysis.

Morphological generalization is often mentioned among the motivations for
character-level modeling. Therefore, we evaluate our models using MorphEval
\citep{burlot-yvon-2017-evaluating,burlot-etal-2018-wmt18}. Similar to the
gender evaluation, MorphEval also uses contrastive sentence pairs that differ
in exactly one morphological feature. Accuracy on the sentences is measured.
Besides, we assess how well the models handle lemmas and forms that were unseen
at training time. We tokenize and lemmatize all data with UDPipe
\citep{straka-strakova-2017-tokenizing}. On the WMT20 test set, we compute the
recall of test lemmas that were not in the training set and the recall of word
forms that were not in the training data, but forms of the same lemma were.
Note that not generating a particular lemma or form is not necessarily an
error. Therefore, we report the recall in contrast with the recall of lemmas
and forms that were represented in the training data.

Character-level models are also supposed to be more robust towards source-side
noise. We evaluate the noise robustness of the systems using synthetic
noise. We use TextFlint \citep{wang-etal-2021-textflint} to generate synthetic
noise in the source text with simulated typos and spelling errors. We generate
20 noisy versions of the WMT20 test set and report the average chrF score.

\subsection{Results}

The main results are presented in Table~\ref{tab:wmtResults}. The main trends
in the translation quality are the same as in the case of IWSLT data: subword
models outperform character models. Using Lee-style encoding narrows the
quality gap and performs similarly to models with subword tokens on the source
side. Although domain robustness often motivates character-level experiments,
our experiments show that the trends are domain-independent, except for
English-German IT Domain translation.

The similar performance of the subword encoder and the Lee-style encoder
suggests that the hidden states of the Lee-style encoder can efficiently
emulate the subword segmentation. We speculate that the main weaknesses remain
on the decoder side.

In the English-to-Czech direction, the character-level models perform worse in
gender bias evaluation, although they better capture grammatical gender
agreement according to the MorphEval benchmark. On the other hand,
character-level models make more frequent errors in the tense of coordinated
verbs.
%
%
There are no major differences in recall of novel forms and lemmas.

For the English-to-German translation, character-level methods reach better
results on the gender benchmark. We speculate that getting gender correct in
German might be easier because unlike Czech it does not require subject-verb
agreement. The average performance on the MorphEval benchmark is also slightly
better for character models. Detailed results on MorphEval are in
Tables~\ref{tab:morphevalCs} and~\ref{tab:morphevalDe} in the Appendix. The
higher recall of novel forms also suggests slightly better morphological
generalization.

The only consistent advantage of the character-level models is their robustness
towards source side noise. Here, the character-level models outperform both the
fully subword model and the subword encoder.

\section{Conclusions}

In our extensive literature survey, we found evidence that character-level
methods should reach comparative translation quality as subword methods,
typically at the expense of much higher computation costs. We speculate that
the computational cost is the reason why virtually none of the recent WMT systems
used character-level methods
or mentioned 
them as a reasonable alternative.

Recently, most innovations in character-level modeling were introduced in the
context of pre-trained representations.
In our comparison of character processing architectures (two of them used for
the first time in the context of MT), we showed that 1D convolutions followed
by highway layers still deliver the best results for MT\@.

Character-level systems are still mostly worse than subword systems. Moreover,
the recent character-level architectures do not show advantages over vanilla
character models, other than improved speed.

To overcome efficiency issues, we proposed a two-step decoding architecture
that matches the speed of subword models, however at the expense of a further
drop in translation quality.

Furthermore, we found that conclusions of recent literature on decoding in MT
do not generalize for character models. Character models do not suffer from the
beam search curse and decoding methods based on sampling perform poorly, here.

Evaluation on competitively large datasets showed that there is still a small
quality gap between character and subword models. Character models do not show
better domain robustness, and only slightly better morphological generalization
in German, although this is often mentioned as important motivation for
character-level modeling. The only clear advantage of character models is high
robustness towards source-side noise.

In contrast to earlier work on character-level MT, which claimed that decoding
is straightforward and which focused on the encoder part of the model, our
conclusions are that Lee-style encoding is comparable to subword encoders.
Even now, most modeling innovations focus on encoding.
Character-level decoding which is both accurate and efficient remains an open
research question.

\section*{Acknowledgement}

Many thanks to Martin Popel for comments on the pre-print of this paper and to
Lukas Edman for discovering a bug in the source code and for a fruitful
discussion on the topic of the paper. \\[1ex]
\noindent The work at LMU Munich was supported by was supported by the European Research
Council (ERC) under the European Union’s Horizon 2020 research and innovation
programme (No.~640550) and by the German Research Foundation (DFG; grant FR
2829/4-1). The work at CUNI was supported by the European Commission via its
Horizon 2020 research and innovation programme (No.~870930).

\bibliographystyle{acl_natbib}
\bibliography{anthology,custom}

\appendix

\section{Two-step decoder}\label{sec:appendixTwoStep}

Here, we describe details of the architecture of the two step decoder shown in
Figure~\ref{fig:TwoStepDecoder}. The input of the decoder are hidden states of
the character processing architecture, i.e., for a downsampling factor $s$, a
sequence that is $s$ times shorter than the input sequence. The output of the
Transformer stack is a sequence of the same length.

For each Transformer decoder state $h_i$, the decoder needs to produce $s$
characters. This is done by a light-weight autoregressive LSTM decoder. In each
step, it has two inputs: the embedding of the previously decoded character and
a projection of the decoder state $h_i$. There are $s$ different linear
projections for each of the output character generated from a single
Transformer state.

At inference time, the LSTM decoder gets one Transformer state and generates
$s$ output characters. The characters are fed to the character processing
architecture, which is in turn used to generate the next Transformer decoder
state.

\section{IWSLT Experiments}\label{appendix:iwslt}

\subsection{Dataset details}

We used the \texttt{tst2010} part of the dataset for validation and
\texttt{tst2015} for testing and did not use any other test sets. The data
sizes are presented in Table~\ref{tab:dataSizes}.

\begin{table*}[t]
    \footnotesize\centering
    \begin{tabular}{l ccc ccc ccc}
        \toprule
        & \multicolumn{3}{c}{Train}
        & \multicolumn{3}{c}{Validation}
        & \multicolumn{3}{c}{Test}

        \\
        \cmidrule(lr){2-4}
        \cmidrule(lr){5-7}
        \cmidrule(lr){8-10}

        & Sent. & \begin{minipage}{1cm}\centering Char.\\src\end{minipage} & \begin{minipage}{1cm}\centering Char.\\tgt\end{minipage}
        & Sent. & \begin{minipage}{1cm}\centering Char.\\src\end{minipage} & \begin{minipage}{1cm}\centering Char.\\tgt\end{minipage}
        & Sent. & \begin{minipage}{1cm}\centering Char.\\src\end{minipage} & \begin{minipage}{1cm}\centering Char.\\tgt\end{minipage}

        \\ \midrule

        en-ar
        & 232k & 22.5M & 32.8M
        & 1.3k & 119k  & 179k
        & 1.2k & 116k  & 164k
        \\

        en-de
        & 206k & 19.9M & 21.7M
        & 1.3k & 117k  & 132k
        & 1.1k & 109k  & 100k
        \\

        en-fr
        & 232k & 22.6M & 25.5M
        & 1.3k & 119k  & 140k
        & 1.2k & 116k  & 129k
        \\

        \bottomrule
    \end{tabular}

    \caption{IWSLT data statistics in terms of number of parallel sentences and
    number of characters.}\label{tab:dataSizes}

\end{table*}

\subsection{Model Hyperparameters}

All models are trained with initial learning rate: $5\cdot 10^{-4}$ with 4k
warmup steps. The batch size is 20k tokens for both BPE and character
experiments with update after 3 batches. Label smoothing is set to 0.1.

\paragraph{Lee-style.} The character embedding dimension is 64. The original
paper used kernel sizes from 1 to 8. For ease of implementation, we only use
even-sized kernels up to size 9. The encoder uses 1D convolutions of kernel
size 1, 3, 5, 7, 9 with 128, 256, 512, 512, 256 filters. Their output is
concatenated and projected to the model dimension, followed by 2 highway layers
and 2 Transformer feed-forward layers.

\paragraph{CANINE.} The local self-attention span in the encoder is 4$\times$
the downsampling factor, in the encoder, equal to the downsampling factor.

\paragraph{Two-step decoder.} The decoder uses character embeddings with
dimension of 64, which is also the size of the projection of the Transformer
decoder state. The hidden state size of the LSTM is 128.

\subsection{Validation Performance}

The validation BLEU and chrF scores and training and inference times are in
Table~\ref{tab:iwsltValidation}. The training times were measured on machines
with GeForce GTX 1080 Ti GPUs and with Intel Xeon E5--2630v4 CPUs (2.20GHz), a single GPU was used.


Note that the experiments on IWSLT were not optimized for speed and are thus
not comparable with the times reported
on the larger datasets.

\begin{table*}

	\newcommand{\BLEU}{B\scalebox{0.5}{LEU}}
\newcommand{\chrF}{chrF}
\newcommand{\R}[2]{\begin{minipage}{21pt}\centering\vspace{2pt} #1 \\ \tiny$\pm$#2\vspace{2pt}\end{minipage}}

\begin{adjustbox}{max width=\textwidth}
\begin{tabular}{llc | cccc|cccc|cccc | cccc|cccc|cccc}
\toprule

\multirow{3}{*}{\rotatebox{90}{Model\hspace{10pt}}} & \multirow{2}{*}{Enc.} &\multirow{2}{*}{Dec.} & 
\multicolumn{12}{c}{From English} & \multicolumn{12}{c}{Into English} \\  \cmidrule(lr){4-15}  \cmidrule(lr){16-27}
& &
& \multicolumn{4}{c}{ar}
& \multicolumn{4}{c}{de}
& \multicolumn{4}{c}{fr}
& \multicolumn{4}{c}{ar}
& \multicolumn{4}{c}{de}
& \multicolumn{4}{c}{fr} \\
\cmidrule(lr){2-3} \cmidrule(lr){4-7} \cmidrule(lr){8-11} \cmidrule(lr){12-15} \cmidrule(lr){16-19} \cmidrule(lr){20-23} \cmidrule(lr){24-27}

& \multicolumn{2}{c}{downsample} &
 Train & Valid & \BLEU & \chrF &
 Train & Valid & \BLEU & \chrF &
 Train & Valid & \BLEU & \chrF &
 Train & Valid & \BLEU & \chrF &
 Train & Valid & \BLEU & \chrF &
 Train & Valid & \BLEU & \chrF \\ \midrule

\multicolumn{3}{l}{BPE 16k}
& \R{8.9}{1.6} & \R{19.4}{1.0} & \R{13.8}{0.2} & \R{.411}{.002}
& \R{8.2}{0.9} & \R{23.8}{8.6} & \R{26.1}{0.3} & \R{.523}{.001}
& \R{6.8}{1.0} & \R{20.6}{1.0} & \R{35.8}{0.3} & \R{.594}{.002}
& \R{10.4}{0.7}& \R{19.8}{0.7} & \R{27.9}{0.1} & \R{.501}{.002}
& \R{8.9}{0.2} & \R{16.2}{1.0} & \R{30.2}{0.1} & \R{.534}{.001}
& \R{9.3}{0.7} & \R{17.4}{0.5} & \R{37.9}{0.3} & \R{.591}{.003}
\\
\multicolumn{3}{l}{Vanilla char.}
& \R{14.5}{5.5} & \R{203.2}{3.9}  & \R{11.4}{0.2} & \R{.417}{.003}
& \R{13.7}{5.5} & \R{293.5}{5.8}  & \R{24.7}{0.5} & \R{.516}{.005}
& \R{17.0}{2.0} & \R{318.7}{3.8}  & \R{34.9}{0.3} & \R{.590}{.002}
& \R{16.2}{5.2} & \R{241.3}{28.1} & \R{26.8}{0.7} & \R{.499}{.005}
& \R{15.6}{3.3} & \R{203.5}{29.6} & \R{29.0}{0.7} & \R{.527}{.004}
& \R{17.9}{2.7} & \R{230.8}{29.4} & \R{36.9}{0.5} & \R{.583}{.003}
\\ \midrule
\multirow{6}{*}{\rotatebox{90}{Lee-style}} & 3 & ---
& \R{13.0}{9.5} & \R{232.8}{3.3} & \R{11.5}{0.1} & \R{.420}{.002}
& \R{16.6}{9.2} & \R{331.0}{7.2} & \R{24.8}{0.1} & \R{.519}{.002}
& \R{11.1}{9.1} & \R{358.2}{7.0} & \R{34.9}{0.4} & \R{.591}{.003}
& \R{9.6}{9.0} & \R{321.0}{1.2} & \R{27.0}{0.1}  & \R{.502}{.002}
& \R{16.5}{8.2} & \R{275.2}{1.1} & \R{29.6}{0.3} & \R{.533}{.003}
& \R{17.4}{7.7} & \R{301.5}{3.4} & \R{37.6}{0.3} & \R{.589}{.002}
\\
 & 5 & ---
& \R{16.5}{6.8} & \R{223.2}{6.9} & \R{11.0}{0.2}  & \R{.411}{.002}
& \R{9.4}{7.4} & \R{313.8}{4.9} & \R{23.6}{0.2}   & \R{.510}{.002}
& \R{18.7}{2.0} & \R{347.5}{3.9} & \R{32.6}{0.4}  & \R{.576}{.002}
& \R{9.2}{7.6} & \R{237.0}{120.7} & \R{23.7}{4.7} & \R{.472}{.043}
& \R{21.3}{1.7} & \R{257.0}{2.9} & \R{28.5}{0.4}  & \R{.524}{.003}
& \R{10.8}{9.6} & \R{287.8}{9.0} & \R{36.4}{0.2}  & \R{.580}{.002}
\\
 & 3 & 3
& \R{15.4}{3.2} & \R{81.5}{2.1} & \R{10.0}{0.2}  & \R{.398}{.002}
& \R{15.7}{3.1} & \R{103.0}{6.0} & \R{22.5}{0.3} & \R{.502}{.002}
& \R{17.1}{2.9} & \R{106.0}{0.7} & \R{33.0}{0.2} & \R{.579}{.000}
& \R{14.2}{8.3} & \R{102.5}{2.2} & \R{24.6}{0.3} & \R{.484}{.001}
& \R{16.2}{2.0} & \R{90.8}{2.9} & \R{27.3}{0.2}  & \R{.513}{.002}
& \R{14.8}{2.2} & \R{94.8}{3.7} & \R{35.3}{0.2}  & \R{.574}{.001}
\\
 & 5 & 5
& \R{13.7}{3.9} & \R{41.0}{0.9} & \R{8.4}{0.1}   & \R{.377}{.002}
& \R{13.1}{5.2} & \R{46.4}{0.8} & \R{19.5}{0.3}  & \R{.474}{.003}
& \R{10.7}{3.4} & \R{44.2}{11.1} & \R{28.0}{1.9} & \R{.545}{.013}
& \R{11.6}{6.8} & \R{47.2}{0.4} & \R{22.1}{0.2}  & \R{.461}{.002}
& \R{10.8}{1.1} & \R{43.4}{0.5} & \R{24.1}{0.2}  & \R{.489}{.002}
& \R{8.9}{2.0} & \R{46.4}{0.8} & \R{31.8}{0.4}   & \R{.549}{.003}
\\ \midrule
\multirow{6}{*}{\rotatebox{90}{Charformer}} & 3 & ---
& \R{16.4}{2.4} & \R{232.0}{8.4} & \R{11.3}{0.2}  & \R{.417}{.002}
& \R{16.4}{2.7} & \R{342.2}{7.1} & \R{24.0}{0.4}  & \R{.510}{.004}
& \R{17.2}{1.5} & \R{363.8}{8.3} & \R{33.7}{0.1}  & \R{.582}{.002}
& \R{15.4}{7.0} & \R{363.0}{40.0} & \R{27.1}{0.3} & \R{.500}{.002}
& \R{16.7}{1.0} & \R{276.0}{4.4} & \R{29.4}{0.3}  & \R{.531}{.001}
& \R{17.9}{3.2} & \R{306.2}{8.3} & \R{37.1}{0.3}  & \R{.587}{.001}
\\
 & 5 & ---
& \R{14.0}{1.9} & \R{63.0}{7.0} & \R{7.4}{0.1}   & \R{.359}{.003}
& \R{12.2}{1.0} & \R{80.8}{15.4} & \R{18.2}{0.2} & \R{.456}{.002}
& \R{13.8}{3.2} & \R{76.2}{7.4} & \R{27.8}{0.5}  & \R{.536}{.005}
& \R{11.5}{3.7} & \R{62.5}{8.0} & \R{18.1}{2.7}  & \R{.419}{.027}
& \R{11.6}{1.4} & \R{64.2}{2.9} & \R{23.0}{0.3}  & \R{.480}{.003}
& \R{13.0}{5.5} & \R{72.5}{9.1} & \R{30.6}{0.3}  & \R{.541}{.002}
\\
 & 3 & 3
& \R{15.5}{1.6} & \R{81.2}{1.5} & \R{10.0}{0.2}  & \R{.398}{.001}
& \R{14.9}{2.3} & \R{102.8}{3.1} & \R{22.5}{0.3} & \R{.497}{.003}
& \R{16.2}{1.1} & \R{119.2}{9.0} & \R{32.2}{0.4} & \R{.571}{.003}
& \R{14.8}{3.8} & \R{104.2}{4.8} & \R{24.8}{0.3} & \R{.482}{.003}
& \R{13.4}{0.7} & \R{89.0}{2.5} & \R{27.6}{0.2}  & \R{.516}{.002}
& \R{15.7}{2.4} & \R{100.2}{9.8} & \R{35.7}{0.1} & \R{.576}{.001}
\\
 & 5 & 5
& \R{14.0}{1.9} & \R{63.0}{7.0} & \R{7.4}{0.1}   & \R{.359}{.003}
& \R{12.2}{1.0} & \R{80.8}{15.4} & \R{18.2}{0.2} & \R{.456}{.002}
& \R{13.8}{3.2} & \R{76.2}{7.4} & \R{27.8}{0.5}  & \R{.536}{.005}
& \R{11.5}{3.7} & \R{62.5}{8.0} & \R{18.1}{2.7}  & \R{.419}{.027}
& \R{11.6}{1.4} & \R{64.2}{2.9} & \R{23.0}{0.3}  & \R{.480}{.003}
& \R{13.0}{5.5} & \R{72.5}{9.1} & \R{30.6}{0.3}  & \R{.541}{.002}
\\ \midrule
\multirow{6}{*}{\rotatebox{90}{Canine}} & 3 & ---
& \R{14.8}{2.2} & \R{300.8}{6.8} & \R{10.7}{0.3}  & \R{.407}{.004}
& \R{19.1}{2.3} & \R{481.0}{51.2} & \R{24.1}{0.2} & \R{.513}{.002}
& \R{20.0}{3.3} & \R{494.8}{13.8} & \R{33.9}{0.6} & \R{.582}{.003}
& \R{19.7}{3.3} & \R{368.8}{3.8} & \R{26.1}{0.3}  & \R{.493}{.003}
& \R{18.5}{2.3} & \R{318.2}{10.2} & \R{28.8}{0.4} & \R{.526}{.003}
& \R{13.3}{6.5} & \R{347.5}{10.1} & \R{36.7}{0.4} & \R{.583}{.003}
\\
 & 5 & ---
& \R{13.9}{7.5} & \R{249.2}{5.0} & \R{9.4}{0.2}   & \R{.386}{.002}
& \R{13.5}{7.3} & \R{366.8}{2.8} & \R{21.6}{0.4}  & \R{.489}{.005}
& \R{20.1}{4.2} & \R{395.5}{5.4} & \R{31.2}{0.7}  & \R{.558}{.005}
& \R{17.7}{4.8} & \R{363.2}{8.9} & \R{22.6}{0.1}  & \R{.458}{.001}
& \R{12.9}{7.5} & \R{300.8}{10.8} & \R{26.7}{0.2} & \R{.508}{.002}
& \R{16.9}{2.5} & \R{312.2}{3.7} & \R{34.4}{0.5} &  \R{.564}{.003}
\\
 & 3 & 3
& \R{17.3}{2.5} & \R{91.5}{1.1}   & \R{9.4}{0.3}  & \R{.390}{.003}
& \R{18.6}{2.8} & \R{138.5}{11.9} & \R{21.6}{0.4} & \R{.493}{.001}
& \R{18.4}{1.8} & \R{132.2}{15.9} & \R{31.6}{0.6} & \R{.567}{.004}
& \R{14.1}{4.9} & \R{115.2}{1.8}  & \R{23.9}{0.6} & \R{.474}{.004}
& \R{12.9}{2.4} & \R{104.5}{4.0}  & \R{26.2}{0.8} & \R{.505}{.006}
& \R{14.2}{5.9} & \R{118.0}{4.1}  & \R{35.0}{0.1} & \R{.572}{.001}
\\
 & 5 & 5
& \R{17.1}{8.3} & \R{72.0}{6.7} & \R{6.1}{0.2}  & \R{.332}{.005}
& \R{15.2}{4.4} & \R{85.5}{9.6} & \R{17.3}{0.3} & \R{.450}{.004}
& \R{16.2}{1.8} & \R{89.0}{5.4} & \R{27.1}{0.3} & \R{.529}{.003}
& \R{20.9}{1.1} & \R{81.8}{1.9} & \R{15.7}{0.4} & \R{.391}{.005}
& \R{15.7}{3.9} & \R{75.0}{2.4} & \R{22.5}{0.2} & \R{.473}{.001}
& \R{13.1}{3.1} & \R{84.5}{5.0} & \R{29.4}{0.1} & \R{.529}{.002}
\\
\bottomrule
\end{tabular}
\end{adjustbox}

	\caption{Training time (hours), inference time on the validation set
	(seconds) and translation quality in terms of BLUE and chrF scores on the
	validation data.}\label{tab:iwsltValidation}

\end{table*}


\section{WMT Experiments}\label{appendix:wmt}

\subsection{Training Details}

We use the Transformer Big architecture as defined FairSeq's standard
\texttt{transformer\_wmt\_en\_de\_big\_t2t}. The Lee-style encoder uses filters
sizes 1, 3, 5, 7, 9 of dimensions 256, 512, 1024, 1024, 512.  The other
parameters remains the same as in the IWSLT experiments.

We set the beta parameters of the Adam optimizer to 0.9 and 0.998 and gradient
clipping to 5. The learning rate is $5\cdot10^{-4}$ with 16k warmup steps.
Early stopping is with respect to negative log likelihood with patience 10. We
save 5 best checkpoints and do checkpoint averaging before evaluation. The
maximum batch size is 1800 tokens for the BPE experiments and 500 for
character-level experiments.  We train the models on 4 GPUs, so the effective
batch size is 4 times bigger.

\subsection{Validation Performance}

During training, we evaluated the models by measuring the cross-entropy on the
validation set. After model training, we use grid search to estimate the best
value of length normalization on 
the
validation set. The translation quality on
the validation data is tabulated in Table~\ref{tab:wmtValidation}.

\begin{table}
    \centering
    \newcommand{\BLEU}{B\scalebox{0.5}{LEU}}
\newcommand{\chrF}{chrF}
\newcommand{\COMET}{C\scalebox{0.4}{OMET}}

\footnotesize
\begin{tabular}{ll ccc ccc}

\toprule

    &    & \BLEU & \chrF & \COMET & \begin{minipage}{1cm}\centering Len. norm.\end{minipage} \\ \midrule

\multirow{4}{*}{\rotatebox{90}{en-cs\phantom{d}}}
    & BPE 16k            & 24.4 & .524 & .753 & 0.8 \\
    & BPE to char        & 22.9 & .513 & .687 & 1.2 \\
    & Vanilla char.      & 22.3 & .506 & .654 & 1.4 \\
    & Lee-style enc.     & 23.1 & .514 & .698 & 1.0 \\

\midrule

\multirow{4}{*}{\rotatebox{90}{en-de}}
    & BPE 16k        & 47.8 & .708 & .651 & 1.2 \\
    & BPE to char    & 43.7 & .683 & .594 & 1.2 \\
    & Vanilla char.  & 42.7 & .675 & .569 & 1.4 \\
    & Lee-style enc. & 43.7 & .684 & .595 & 1.6 \\

\bottomrule

\end{tabular}

    \caption{Translation quality on the validation data and the value of length
    normalization that led to the best quality.
    }\label{tab:wmtValidation}

\end{table}

\subsection{Detailed Results}

The detailed results on the MorphEval benchmark are in
Tables~\ref{tab:morphevalCs} (Czech) and~\ref{tab:morphevalDe} (German). The
details of the noise evaluation are in Table~\ref{tab:noise}.

\begin{table}

    \centering
    \begin{adjustbox}{max width=\columnwidth}
\footnotesize
\begin{tabular}{l cccc}
    \toprule
 & BPE & BPE2char & char & lee \\ \midrule

comparative & 78.2\% & 78.2\% & 79.6\% & 80.4\% \\
conditional & 59.8\% & 65.8\% & 71.2\% & 68.4\% \\
coordverb-number & 85.4\% & 81.2\% & 77.4\% & 80.0\% \\
coordverb-person & 85.2\% & 82.0\% & 78.0\% & 80.0\% \\
coordverb-tense & 81.8\% & 78.4\% & 74.0\% & 75.2\% \\
coref-gender & 71.7\% & 74.8\% & 76.5\% & 75.9\% \\
future & 86.2\% & 85.8\% & 84.0\% & 85.8\% \\
negation & 96.2\% & 97.4\% & 98.0\% & 98.2\% \\
noun number & 79.4\% & 81.0\% & 80.8\% & 81.4\% \\
past & 87.2\% & 89.0\% & 89.4\% & 86.8\% \\
preposition & 96.0\% & 96.6\% & 96.1\% & 95.9\% \\
pron2coord & 100.0\% & 100.0\% & 99.6\% & 100.0\% \\
pron2nouns-case & 95.8\% & 95.6\% & 94.4\% & 94.6\% \\
pron2nouns-gender & 95.2\% & 95.2\% & 93.6\% & 93.8\% \\
pron2nouns-number & 95.6\% & 95.6\% & 94.4\% & 94.6\% \\
pron fem & 94.0\% & 94.6\% & 93.8\% & 93.2\% \\
pron plur & 92.0\% & 92.0\% & 92.0\% & 91.4\% \\
pron relative-gender & 78.9\% & 81.8\% & 81.8\% & 81.5\% \\
pron relative-number & 80.1\% & 83.1\% & 82.8\% & 82.6\% \\
superlative & 93.0\% & 91.4\% & 91.0\% & 92.0\% \\ \midrule
NOUN case & .102 & .108 & .105 & .100 \\
ADJ gender & .198 & .194 & .211 & .202 \\
ADJ number & .198 & .190 & .213 & .202 \\
ADJ case & .204 & .198 & .220 & .207 \\
VERB number & .117 & .103 & .101 & .104 \\
VERB person & .091 & .083 & .085 & .084 \\
VERB tense & .113 & .109 & .108 & .110 \\
VERB negation & .081 & .077 & .075 & .075 \\ \midrule
Average & 88.6\% & 87.0\% & 86.4\% & 86.6\& \\

    \bottomrule
\end{tabular}
\end{adjustbox}

    \caption{Detailed MorphEval results for English-Czech
    translation.}\label{tab:morphevalCs}

\end{table}

\begin{table}

    \centering
    \begin{adjustbox}{max width=\columnwidth}
\footnotesize
\begin{tabular}{l cccc}
    \toprule
 & BPE & BPE2char & Char & Lee \\ \midrule

dj strong & 97.9\% & 98.7\% & 99.6\% & 99.2\% \\
comparative & 96.9\% & 96.8\% & 95.6\% & 96.3\% \\
compounds syns & 65.9\% & 66.0\% & 65.4\% & 66.7\% \\
conditional & 90.5\% & 95.4\% & 97.0\% & 97.0\% \\
coordverb-number & 98.0\% & 98.7\% & 99.1\% & 99.3\% \\
coordverb-person & 98.3\% & 99.1\% & 99.5\% & 99.8\% \\
coordverb-tense & 98.0\% & 98.7\% & 99.3\% & 99.3\% \\
coref-gender & 94.5\% & 93.2\% & 95.1\% & 91.9\% \\
future & 87.3\% & 90.8\% & 87.6\% & 88.9\% \\
negation & 98.8\% & 98.8\% & 99.4\% & 99.4\% \\
noun number & 67.0\% & 69.3\% & 71.5\% & 68.4\% \\
past & 94.7\% & 97.1\% & 96.0\% & 96.5\% \\
pron2nouns-gender & 100.0\% & 100.0\% & 100.0\% & 100.0\% \\
pron2nouns-number & 100.0\% & 100.0\% & 100.0\% & 100.0\% \\
pron plur & 99.2\% & 99.2\% & 98.6\% & 98.2\% \\
pron relative-gender & 69.4\% & 69.1\% & 68.8\% & 71.0\% \\
pron relative-number & 69.4\% & 69.1\% & 68.8\% & 71.0\% \\
superlative & 99.8\% & 99.8\% & 99.8\% & 99.6\% \\
verb position & 96.0\% & 95.2\% & 95.2\% & 95.8\% \\ \midrule
ADJ gender & .006 & .002 & .002 & .003 \\
ADJ number & .004 & .001 & .002 & .001 \\
NOUN case & .018 & .011 & .013 & .011 \\
VERB number & .022 & .017 & .015 & .020 \\
VERB person & .010 & .010 & .006 & .008 \\
VERB tense/mode & .046 & .041 & .049 & .050 \\ \midrule
Average & 90.6 & 91.3 & 91.4 & 91.5 \\

\bottomrule
\end{tabular}
\end{adjustbox}

    \caption{Detailed MorphEval results for English-German
    translation.}\label{tab:morphevalDe}

\end{table}

\begin{table}

    \centering
    \newcommand{\BLEU}{B\scalebox{0.5}{LEU}}
\newcommand{\chrF}{chrF}
\newcommand{\COMET}{C\scalebox{0.4}{OMET}}

\newcommand{\R}[2]{#1 \tiny$\pm$#2}

\footnotesize
\begin{tabular}{ll ccc}

\toprule

    &    & BLEU & chrF & COMET \\ \midrule

\multirow{4}{*}{\rotatebox{90}{en-cs\phantom{d}}}
    & BPE 16k            & \R{15.1}{0.2} & \R{.436}{.002} & \R{-.863}{.010} \\
    & BPE to char        & \R{14.4}{0.2} & \R{.436}{.001} & \R{-.836}{.009} \\
    & Vanilla char.      & \R{19.5}{0.2} & \R{.493}{.001} & \R{-.307}{.009} \\
    & Lee-style enc.     & \R{20.2}{0.2} & \R{.497}{.001} & \R{-.308}{.009} \\

\midrule

\multirow{4}{*}{\rotatebox{90}{en-de}}
    & BPE 16k        & \R{16.0}{0.2} & \R{.464}{.002} & \R{-1.127}{.012} \\
    & BPE to char    & \R{15.5}{0.2} & \R{.465}{.001} & \R{-1.112}{.008} \\
    & Vanilla char.  & \R{18.5}{0.1} & \R{.504}{.001} & \R{-.742}{.013} \\
    & Lee-style enc. & \R{19.6}{0.1} & \R{.515}{.001} & \R{-.743}{.014} \\

\bottomrule

\end{tabular}

    \caption{Detailed results on the datasets with generated noise. Average and
    standard deviation for 20 evaluations.}\label{tab:noise}

\end{table}

\end{document}